\documentclass[10pt,twocolumn,letterpaper]{article}

\usepackage{iccv}
\usepackage{times}
\usepackage{epsfig}
\usepackage{graphicx}
\usepackage{amsmath}
\usepackage{amssymb}

\usepackage{multirow}
\usepackage{makecell}
\usepackage{booktabs}
\usepackage{hhline}

\usepackage{float}

\usepackage[breaklinks=true,bookmarks=false]{hyperref}

\iccvfinalcopy 

\def\iccvPaperID{****} 

\begin{document}

\title{Object-Driven Multi-Layer Scene Decomposition From a Single Image}

\author{Helisa Dhamo $^{1}$\\
{\tt\small helisa.dhamo@tum.de}\\
\and
Nassir Navab $^{1}$\\
{\tt\small nassir.navab@tum.de}
\and
Federico Tombari $^{1,2}$\\
{\tt\small tombari@in.tum.de} \\
\and
$^{1}$ Technische Universit\"at M\"unchen
\and
$^{2}$ Google 
}


\maketitle


\begin{abstract}
    We present a method that tackles the challenge of predicting color and depth behind the visible content of an image. Our approach aims at building up a Layered Depth Image (LDI) from a single RGB input, which is an efficient representation that arranges the scene in layers, including originally occluded regions. Unlike previous work, we enable an adaptive scheme for the number of layers and incorporate semantic encoding for better hallucination of partly occluded objects. Additionally, our approach is object-driven, which especially boosts the accuracy for the occluded intermediate objects.
    The framework consists of two steps. First, we individually complete each object in terms of color and depth, while estimating the scene layout. Second, we rebuild the scene based on the regressed layers and enforce the recomposed image to resemble the structure of the original input. The learned representation enables various applications, such as 3D photography and diminished reality, all from a single RGB image. \footnote{Project page: \url{https://he-dhamo.github.io/OMLD/}}

\end{abstract}
\section{Introduction}
Completing a scene beyond the partial occlusion of its components is a strongly desired property for many computer vision applications. 
For instance, in robotic manipulation, the ability to see the full target object despite the presence of occluding elements can lead to a more successful and precise grasping. 
In the autonomous driving context the estimation of the full profile and location of potential obstacles occluded by the car in front of us would prove useful to increase the robustness of the trajectory planning and safety control.  

Scene completion beyond occlusion is important not just to improve higher-level perception systems, but also to enhance the fruition of captured visual data. 3D photography uses image content behind occlusion to enhance the user experience while looking at a photo by synthesizing novel unseen views.
When changing the vantage point the picture was originally taken from, the visual content around object borders gets dis-occluded, thus enabling the image fruition to become more immersive. The combination of 3D photography with a virtual reality (VR) display, such as a Head Mounted Display (HMD), holds the potential to generate a visually effective application. 

\begin{figure}[!t]
\centering
\includegraphics[width=0.92\linewidth]{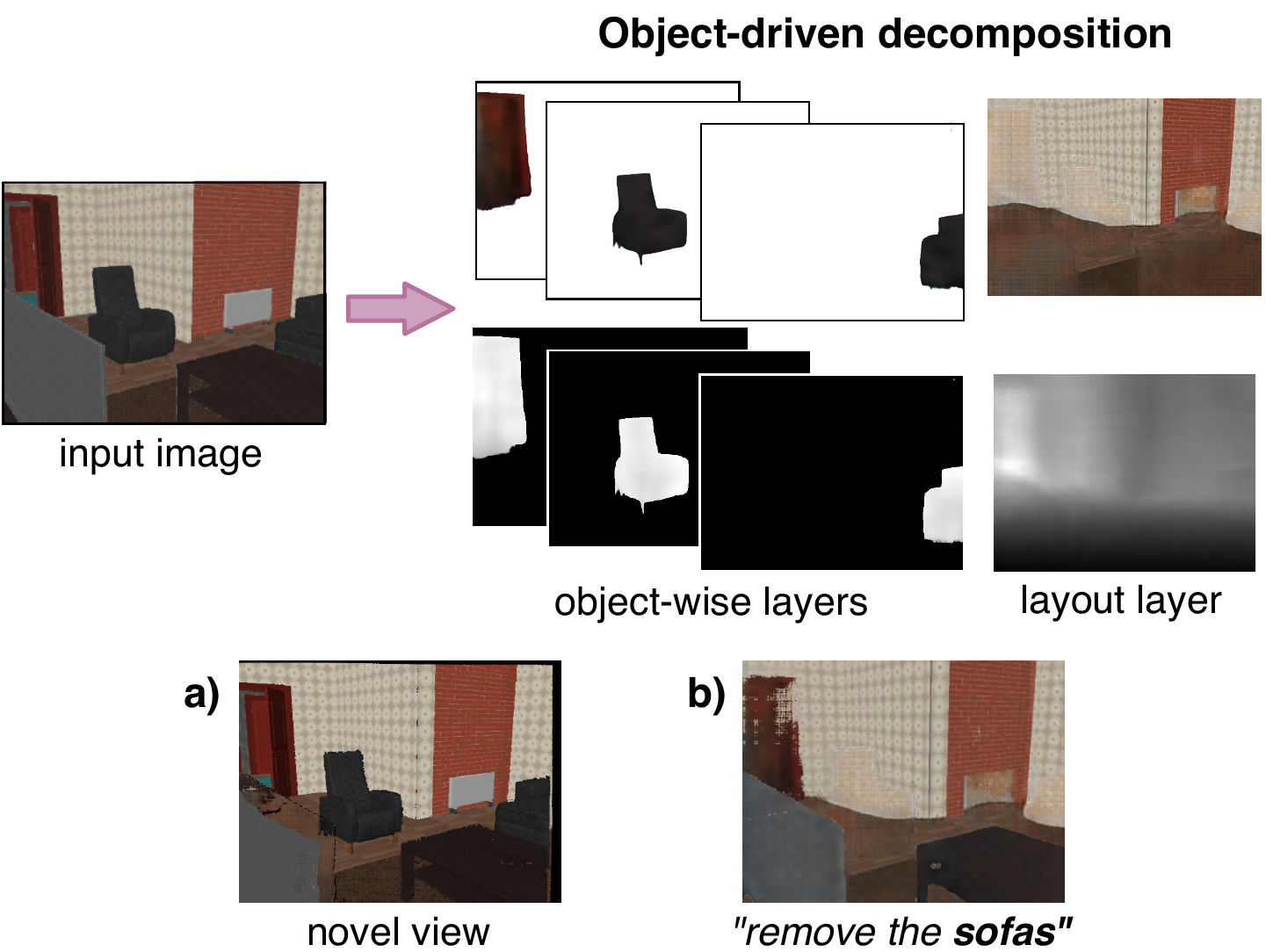}
\caption{\textbf{Overview of our method and its applications.} \emph{(Top):} Given a single color image, we infer a layered representation that consists of a set of RGB and depth images for every object in the scene, as well as the scene layout. \emph{(Bottom):} Illustration of two applications, \textbf{a)} view synthesis and \textbf{b)} object removal.}
\label{fig:teaser}
\end{figure}

Layered Depth Image (LDI), pioneered by Shade \etal \cite{Shade:1998:LDI:280814.280882}, is a data representation distinctively suitable for the aforementioned applications. To augment a single view into a 3D photo, a single depth image is not enough, since it is not designed to store visual and geometric information beyond that of the visible object parts in the scene. 
On the other hand, having a fully completed 3D model of the scene is often an unnecessary complication, since most of the information present in such model would never be used if the novel vantage points are either nearby the original one and/or small in number. It is worth noting that generating such completed 3D scenes typically comes with high computational and memory cost \cite{Zou2018LayoutNetRT,Guo2014PredictingC3,song2016ssc,wang2018}.

Therefore, a layered structure of the original view represents an interesting trade-off between simplicity in the representation and capacity of storing all necessary information for these application. 
Recent works generate such a data representation either from multi-view \cite{Hedman:2017:CP:3130800.3130828}, or stereo \cite{Zhou:2018:SML:3197517.3201323} input, with some variations in the representation. More adventurous approaches \cite{Dhamo2018PeekingBO, lsiTulsiani18} aim to learn an LDI from a single color image. The motivation is providing a method that does not rely on the availability of appropriate photo pairs/sets, so that consumers can reconstruct a 3D photo out of any image, casually at hand. Both Dhamo \etal \cite{Dhamo2018PeekingBO} and Tulsiani \etal \cite{lsiTulsiani18} report results with LDIs having two layers only (background and foreground).

Intuitively, we are able to guess the complete appearance of the partially occluded objects we see, using the color information from the visible parts, together with some object specific characteristics. Here, we motivate a similar feature learning process in our proposed framework. Given the accessibility of state-of-the-art object detectors \eg Mask R-CNN \cite{He2017MaskR}, we assume that predicted instance masks (partial visibility map) and class categories are available. The proposed approach is based on object-wise RGB-D completion followed by a re-composition branch which we call \emph{minimum depth pooling}, that inspects the reconstruction of the original image from the layered representation. This improves the depth prediction accuracy, in that it aligns the visible parts to obtain a global consistency. The proposed method uses the predicted mask probabilities as an attention guiding for the appearance of every object, while the class category predictions aim to induce priors for the object completion. 

Our work aims to bridge current limitations in LDI prediction from a single image, and relies on four main contributions. First, we propose a flexible extension for a multiple layer representation, where unlike \cite{lsiTulsiani18}, the number of layers is not pre-defined in the form of CNN architecture branches. Second, we leverage predictions of semantic identities to perform object-oriented completions, which are not considered in the generation procedure of previous methods \cite{Dhamo2018PeekingBO, lsiTulsiani18}. Third, we propose a re-composition loss that is specific to our task. As a fourth contribution, we generate two datasets suitable for learning layered representations, which we will publicly release for further research. 
 
We show that our results outperform state-of-the-art methods in LDI prediction and view synthesis. In addition, along with view synthesis, our object-level separation enables a new application, the removal of particular target objects, in a diminished reality scenario, Fig. \ref{fig:teaser}. 

\begin{figure*}[t]
\centering
\includegraphics[width=1.03\textwidth]{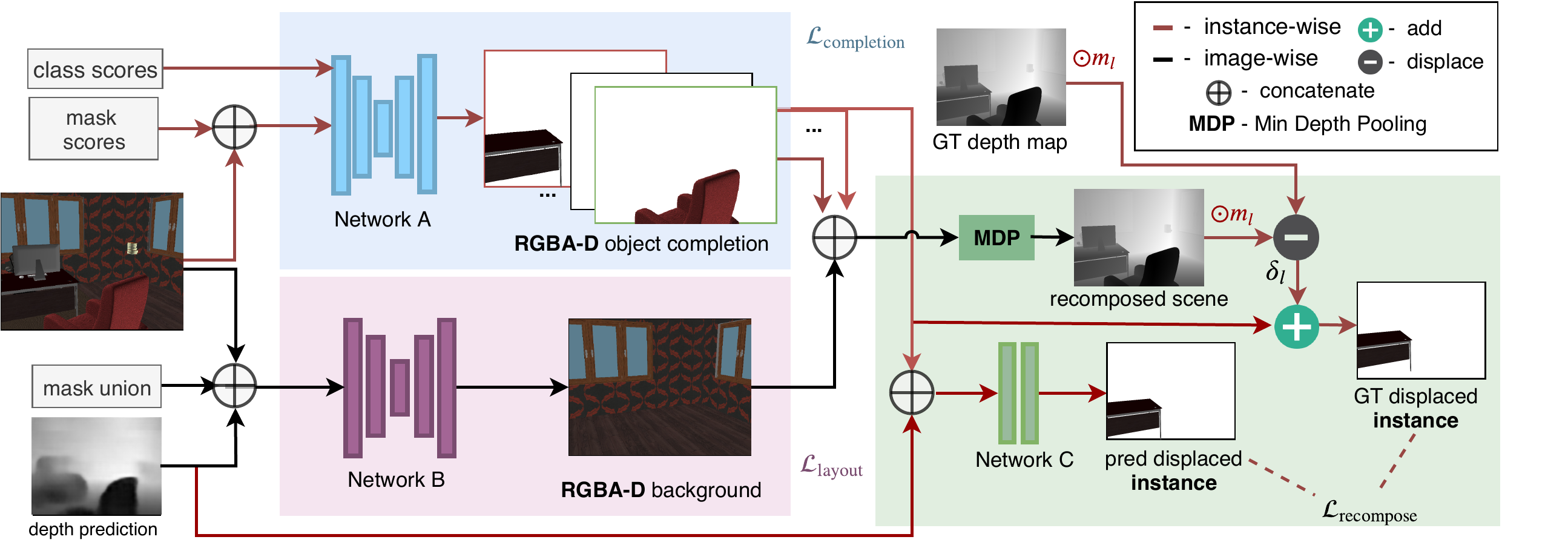}
\caption{\textbf{Proposed scene layering framework.} \emph{Left:} While \emph{Network A} (\emph{top}) completes the occluded parts for each detected object instance, to a RGBA-D representation, \emph{Network B} (\emph{bottom}) predicts RGBA and depth images for the empty scene. \emph{Right:} The outputs are concatenated and fed to the Minimum Depth Pooling (MDP) layer, that recomposes the scene. Instance-wise, the displacement of the recomposed first layer depth from the ground truth depth is used in the re-composition loss to supervise \emph{Network C} and give the final result.}
\label{fig:method}
\end{figure*}
\section{Related Work}

Recent developments in computer vision, extensively target the inference of 3D content from monocular 2D images, either in 2.5D (depth map, normal), object/scene 3D models and layered representations.  

Depth prediction from single view is widely tackled with CNNs, excluding the first few works that consisted of hand-engineered features \cite{saxena2006learning,saxena2009make3d} and data-driven methods \cite{konrad20122d,karsch2016depth}. Eigen et al. \cite{Eigen:2014:DMP:2969033.2969091} propose a multi-scale CNN architecture. Roy and Todorovic \cite{Roy2016MonocularDE} propose regression forests with a shallow CNN on each node. Deeper fully convolutional architectures \cite{laina2016deeper,kendall2017uncertainties} were later proposed based on ResNet \cite{DBLP:journals/corr/HeZRS15} and DenseNet \cite{huang2017densely} respectively. Commonly, Conditional Random Fields (CRFs) \cite{li15,Liu14,Wang_2015_CVPR,xu,LiuSL14} are used to enforce geometrical constraints. Other works exploit semantics to further boost depth performance \cite{Liu2010SingleID,Jiao2018LookDI}.

CNNs have been also applied to 3D inference from a single color image. A family of methods restrict the output to a single object 3D model \cite{choy20163d,Fan_2017_CVPR,3dgan,marrnet}. In contrast, Tulsiani \etal \cite{factored3dTulsiani17} infer a factored 3D scene model composed of a layout and a set of object shapes. None of these methods predicts texture behind occlusion, which is subject of our approach. Other methods exploit more extended inputs to predict 3D scene representations, such as a panorama image \cite{Zou2018LayoutNetRT}, RGB-D \cite{Guo2014PredictingC3} or a depth map \cite{song2016ssc,wang2018}.

Layered scene representations come in a diversity of contexts, such as depth ordering of semantic maps \cite{Yang:2012:LOM:2360767.2361219, Tighe2014ScenePW,Isola2013SceneCA} and color images \cite{ehsani2018segan}, motion analysis and optical flow \cite{moving-layers,Sun:CVPR:2012}, stereo reconstruction \cite{layered-stereo}, scene decomposition in depth surfaces \cite{liu2016layered} and planes \cite{Liu_2019_CVPR}. Our focus is on the Layered Depth Images (LDI) introduced by Shade \etal \cite{Shade:1998:LDI:280814.280882}, which refer to a single view representation of a scene that contains multiple layers of RGB-D information. This can be used for efficient image-based rendering on view perturbations, to deal with information holes on dis-occlusion. Hedman \etal \cite{Hedman:2017:CP:3130800.3130828} use such a representation for 3D photo capturing from multi-view inputs. Zhou \etal \cite{Zhou:2018:SML:3197517.3201323} infer a similar representation from stereo input, that decomposes an image into sweep planes with fixed depth. Very recent works, \cite{Dhamo2018PeekingBO, lsiTulsiani18} propose LDI prediction from a single RGB image, which has the practical advantage of enabling the 3D enhancement of any photo, even if additional views or depth maps are not available. Dhamo \etal \cite{Dhamo2018PeekingBO} automatically generate ground truth data from large-scale datasets that contain trajectory poses, by warping multiple frames to a target view, to populate the second layer of the target LDI. Then, LDIs are learned in a supervised way, formulated as depth prediction and RGB-D inpainting. Tulsiani \etal \cite{lsiTulsiani18} instead, overpass the data limitations by proposing a view synthesis supervision, where the LDIs are learned in a self-supervised way.

\section{Method}

In this Section we detail the proposed multi-layer scene decomposition method. Section \ref{sec:data} presents the rendering procedure we employ to generate ground truth data. Then, the following Sections describe the learning model, which consists of three components: object completion, layout prediction, and image re-composition. These three stages are reported in Fig \ref{fig:method}, that sketches the overall architecture of our model. Since scenes come with varying levels of complexity, assuming a fixed, pre-determined number of layers to represent them tends to limit the flexibility and, as such, the performance of image decomposition approaches. We introduce an adaptive model where the number of layers is dependent on the number of the detected object instances in the current scene.

\subsection{Data generation}
\label{sec:data}

LDI prediction from a single image is quite a novel task, therefore large-scale datasets suitable for deep learning are not available. To overpass this limitation, one could either formulate an indirect supervision \cite{lsiTulsiani18}, or investigate ways to generate ground truth layered image representations \cite{Dhamo2018PeekingBO}. In this work we explore the latter, to investigate the potential advantage of a rich supervision. Our goal of object-oriented layer inference implies the need for additional ground truth, such as RGBA-D representations of every object and layout of the scene, which we acquire automatically from existing datasets. Unlike Dhamo \etal \cite{Dhamo2018PeekingBO}, we employ a mesh-based rendering approach. The advantage is that the 3D mesh captures all the available information in the scene, while an image-based approach \cite{Dhamo2018PeekingBO} only captures information which is present in the set of consecutive image frames to be warped. For every frame, we render the visible instances separately, similarly to \cite{ehsani2018segan}. In addition to the color images and the visibility masks, we extract depth maps and object categories for every instance. For this purpose, we utilize instance annotations, which are available in the 3D meshes, to separately extract the vertices that belong to each visible object in every view frame. Structural elements, identified by their semantic category (floor, walls, ceiling, window) are grouped together in the layout layer. We make sure that instances that were not originally visible in a certain view, are not included in its layered representation. For example, we do not want the object from another room (behind the enclosing wall of the currently visible room) to form part in the compositional layers of that view. The advantage of the proposed semantic-aware rendering with respect to \cite{lsiTulsiani18,Dhamo2018PeekingBO} is that it enables learning of class specific features, which might turn helpful in regressing plausible objects in the novel LDI layers.

Here, we work with two different datasets, namely SunCG \cite{song2016ssc} and Stanford 2D-3D \cite{2017arXiv170201105A_armeni}. Both datasets contain scene meshes together with 2D modalities (color, depth, instances, semantics). The latter suffers from a typical real-world mesh nuisance, namely the presence of holes and missing surface parts, which is critical in our task as it leads to incomplete object renderings behind occlusion. However, we observed in Stanford 2D-3D \cite{2017arXiv170201105A_armeni} fewer such holes compared to other state-of-the-art large-scale real datasets. Through a post-processing step, we select a subset of the rendered layered images, \ie all those where the amount of overlap between layers is beyond a threshold, to ensure that there is enough novel information on dis-occlusion. We use SunCG \cite{song2016ssc} for a fair comparison and ablation, given pixel-perfect ground truth, while Standford 2D-3D \cite{2017arXiv170201105A_armeni} demonstrates applicability in a real world setup.

The generated object and layout layers can be easily arranged in an LDI representation, using the depth maps to sort the layers at every pixel location.

\subsection{Object Completion}
\label{sec:completion}

The goal of the object completion branch (Network A, Fig. \ref{fig:method}) is to learn a mapping from an occluded object $x_c$ to the RGB-D representation of it in full visibility $y$. We use a helping mask for each object, as an intuitive prior for ambiguous problems \cite{Fan_2017_CVPR,ehsani2018segan,Dhamo2018PeekingBO}. In addition, we incorporate semantic classes so to encourage the model to learn class specific properties. Therefore, Network A receives the input RGB image, the predicted mask and class scores of an object, and predicts a five-channel output -- \ie, the completed RGBA-D representation of that object. The algorithm is applied instance-wise, for each available mask. The architecture details are provided in Section \ref{sec:details}. For this task, we found it more adequate to feed full images in Network A instead of object-focused regions of interest (RoIs). Although RoI cropping is more efficient, it limits the generation to a fixed resolution, which mostly affects the texture details of big objects. In addition, it weakens depth perception, as it reduces global context and hinders the understanding for object scaling and extent.

During training, for each ground truth instance mask we determine the respective prediction from Mask R-CNN \cite{He2017MaskR}. The match is defined as the highest intersection-over-union (IoU) between the predicted and ground truth masks. To avoid wrong correspondences, we discard matches that have an $IoU<0.3$. Utilizing these valid ground truth - prediction pairs, we learn how to complete the occluded part of the objects in the image. We use an L1 loss on the RGBA-D predictions $\hat{y}$, which is weighted by a relevance map $\gamma$
\begin{equation}
    \mathcal{L}_\mathrm{completion} = ||\gamma (y - \hat{y})||_1.
    \label{eq:completion}
\end{equation}
The need of $\gamma$ arises from the fact that the different regions in the image have different relevance for the object completion problem. We want to set a higher influence of the loss in pixels close to the object appearance. First, we set $\gamma=0.7$ in the \emph{visible area} of the object in the original image including \emph{a close neighbourhood} (by applying a $31\times31$ dilation in the ground truth mask), $\gamma=1.5$ in the \emph{occluded regions} of that object, and $\gamma=0.2$ otherwise. 
During inference, each mask and category prediction provided by Mask R-CNN \cite{He2017MaskR} is fed to the object-completion network together with the input color image.

Object completion from a single image is a challenging problem, in that predicting the visible part only might represent already a relatively good solution and the occluded parts are not properly learned. The guiding masks are not pixel perfect, therefore the model has to learn to differentiate between the different object instances along the edges, concurrently with the original goal of learning plausible object completions. Therefore, as additional pre-training stage to our learning strategy, we aim to encode common object properties to help our object completion network generate plausible outputs. In this pre-training stage, we train Network A as an auto-encoder (\ie, in an unsupervised way) on single RGBA-D object representations
\begin{equation}
    \mathcal{L}_\mathrm{auto} = ||x - \hat{x}||_1,
\end{equation}
where $x$ is the ground truth RGBA-D map and $\hat{x}$ the auto-encoded output. For this unsupervised learning part, we use those object instances that do not touch the image borders to guarantee visibility on the whole object appearance. Then, we freeze the decoder (including the bottleneck layer) and train the encoder for object completion using the supervised strategy described in eq. (\ref{eq:completion}). Intuitively, this encourages the occluded objects to share the same latent space as their respective RGBA-D representation, in full visibility.

\begin{figure}[!t]
\centering
\includegraphics[width=0.99\linewidth]{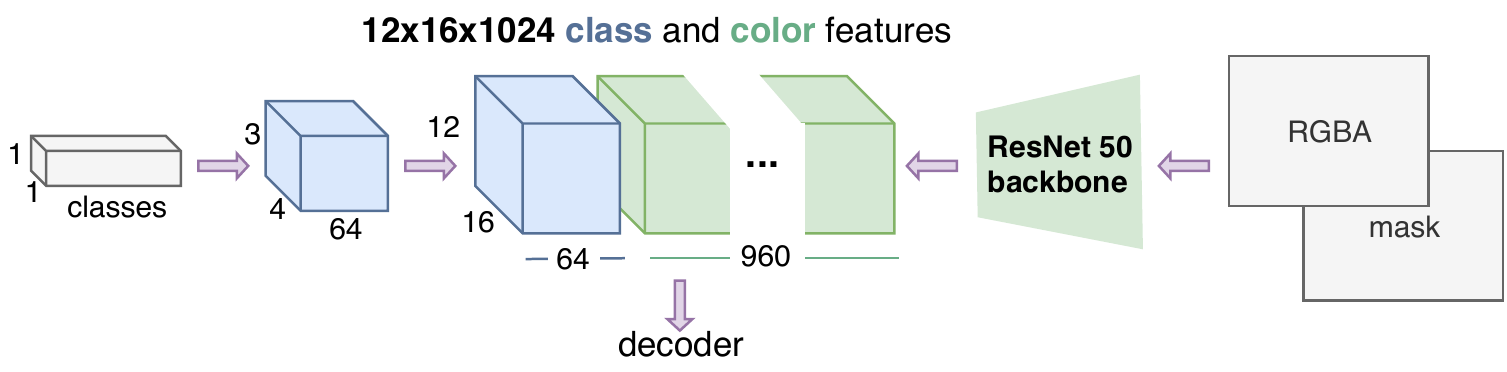}
\caption{\textbf{Overview of the proposed object completion encoder architecture.} Class probabilities branch \emph{(left)} and image branch \emph{(right)} are concatenated along channels in the bottleneck layer.}
\label{fig:details_object}
\end{figure}

\subsection{Layout Prediction}
\label{sec:layout}

The layout branch (Network B in Fig. \ref{fig:method}) is designed to find a mapping between the input RGB image $x_c$ and the corresponding RGB-D scene layout $y$, \ie the object-free representation of the scene.

We employ a fully-convolutional network with skip connections as in \cite{unet}, whose details are provided in Section \ref{sec:details}. We observed considerable improvement on the depth layout prediction, when a standard depth map is regressed before-hand and provided as prior to the layout network. In contrast, such input hindered the optimization performance on the object branch, which we relate with the blurred and inaccurate object edges on such depth maps.  Hence, our model receives an RGBA image $x_c$, a depth map $\hat{x}_d$ predicted via a CNN (in our experiments we use \cite{laina2016deeper}), as well as the union of all the predicted instance masks used in Section \ref{sec:completion}. The mask union is used to give the model a hint on where the instance-free regions are located, so that it could exploit them to extrapolate layout features. Unlike the background inpainting in \cite{Dhamo2018PeekingBO}, we do not mask out the non-structural regions of the image. We wish to point out that, since the masks are simply predictions and not ground truth annotations, they are noisy and can mask out useful content from the image.

With the goal of predicting visually appealing layouts, we propose to carry out this task by means of an adversarial approach \cite{Goodfellow:2014:GAN:2969033.2969125}. In addition, we want to encourage edge consistency in our generations, as the room contour is an interesting property of the layout. A similar motivation has been explored in \cite{zhang2018deepdepth}, where occlusion boundaries are predicted as an intermediate step to improve a depth completion task. Unlike \cite{zhang2018deepdepth}, instead of explicitly generating and supervising edges, we incorporate the perceptual loss $\mathcal{L}_p$ from \cite{Johnson2016PerceptualLF}. Both the generated and ground truth layouts are fed in a VGG-16 \cite{vgg14c} network, pre-trained on ImageNet \cite{imagenet_cvpr09}. Then, in addition to the standard reconstruction loss $\mathcal{L}_r$, we want to exploit the $\mathcal{L}_1$ distance between the respective feature maps at a certain layer of the VGG-16 network. We choose to extract the features from the first VGG block, as we observed that it captures edges as desired. Then, the complete optimization problem becomes
\begin{equation}
    \mathcal{L}_\mathrm{layout} = \lambda_r \mathcal{L}_r + \lambda_p \mathcal{L}_p + \min\limits_G \max\limits_D(\mathcal{L}_\mathrm{a})
\end{equation}
\begin{equation}
    \mathcal{L}_\mathrm{r} = ||y_c - \hat{y}_c||_1 + ||y_d - \hat{y}_d||_1
\end{equation}
\begin{equation}
    \mathcal{L}_\mathrm{p} = ||\phi(y_\mathrm{c}) - \phi(\hat{y}_\mathrm{c})||_1
\end{equation}
\begin{equation}
    \mathcal{L}_\mathrm{a} = \mathbb{E}_{x_c,y_c}[log D(x_{c},y_{c})] + 
\mathbb{E}_{x_c}[log(1-D(x_{c}, G(x_{c})))],
\end{equation}

\noindent where $y_c$, $y_d$ denote output color and depth respectively, and $\phi$ is the output feature map of the first VGG-16 block. Our layout prediction shares the definition and motivations of the one proposed in \cite{factored3dTulsiani17}, however they differ in two aspects. First, we provide a background mask for attention guiding, and second, we regress an additional texture component.

 A full ablation on the improvement of our design choices is provided as supplementary material.

\subsection{Image Re-composition}

An important requirement for our method is to regress layers that are correctly sorted in depth. This means, \eg in the case of a scene with a foreground table in front of a background wall, that the distance between the viewpoint and the wall should not be smaller than the distance between the same viewpoint and the table. To implicitly and globally enforce the depth consistency of all regressed scene parts and layout, we propose an additional component in our model that we dub \emph{minimum depth pooling} (MDP). This concatenates all the predicted layers (including objects and layout) and, for every pixel, extracts the RGBA-D from the layer with the lowest depth. The result of MDP is in the best case identical to the original input and the corresponding visible depth, together with an index map $i_{map}$ that is the argmin of depth.
This \emph{image re-composition} strategy enforces the predicted multi-layer representation to coherently encode the structure of the original input image.

Since depth predictors learn a global understanding for depth, we use a standard predicted depth map $\hat{x}_d$ as a prior for our layer sorting problem. Foe each layer $l$, we only keep the region in $\hat{x}_d$ in which $l$ is the front structure, using a binary mask $m_l$, which is one if $i_{map} = l$, and zero otherwise. We incorporate a re-composition block (\emph{Network C}), that receives the predicted instance depth $\hat{y}_{d,l}$, concatenated with the masked $\hat{x}_d$ from [22]. Then we learn a set of depth displacements $\delta_l$, as the distance between the mean ground truth $y_{d,l}$ and the mean predicted $\hat{y}_{d,l}$ layer depth (over $m_l$)
\begin{equation}
    \delta_{l} = \frac{\sum m_l \odot y_{d,l}}{\sum m_l} - \frac{\sum m_l \odot \hat{y}_{d,l}}{\sum m_l} ,
\end{equation}

\begin{equation}
    \mathcal{L}_\mathrm{recompose} = ||y_{\delta,l} - \hat{y}_{d,l}||_1
\end{equation}

\noindent where $\odot$ is the element-wise multiplication and $y_{\delta,l} = \hat{y}_{d,l} + \delta_l$ is the displaced depth for instance $l$.

We show in experiments that the proposed MDP-driven re-composition loss improves not only the visible region of the objects, but also the occluded parts, given that it allows the whole layer to shift towards the right direction.

\section{Implementation details}
\label{sec:details}

In this section, we provide a more detailed description regarding our architecture choices and implementation.

\textbf{Network A} The object completion network receives two inputs, the first one being an RGBA image concatenated with mask confidences, and the second a vector of class scores, both predicted from Mask R-CNN \cite{He2017MaskR}, Fig. \ref{fig:details_object}. The images are fed into a ResNet-50 \cite{DBLP:journals/corr/HeZRS15} backbone, with the original fully connected layer removed. We append one more convolution layer with $\mathrm{out}_\mathrm{channels}=960$. The second path consists of two deconvolution layers of 64 channels applied consecutively on the class probabilities, which is a feature vector whose size equals the number of classes. Both branch outputs are concatenated along the channels, followed by \emph{layer normalization} \cite{DBLP:journals/corr/BaKH16_layernorm}. The network decoder consists of five up-projection layers from \cite{laina2016deeper}.
The architectures of the auto-encoder and the object completion network are identical, however, since the input modalities are partially different, we learn the encoder from scratch instead of fine-tuning the autoencoder weights. 

\begin{figure*}[t]
\centering
\includegraphics[width=0.9\linewidth]{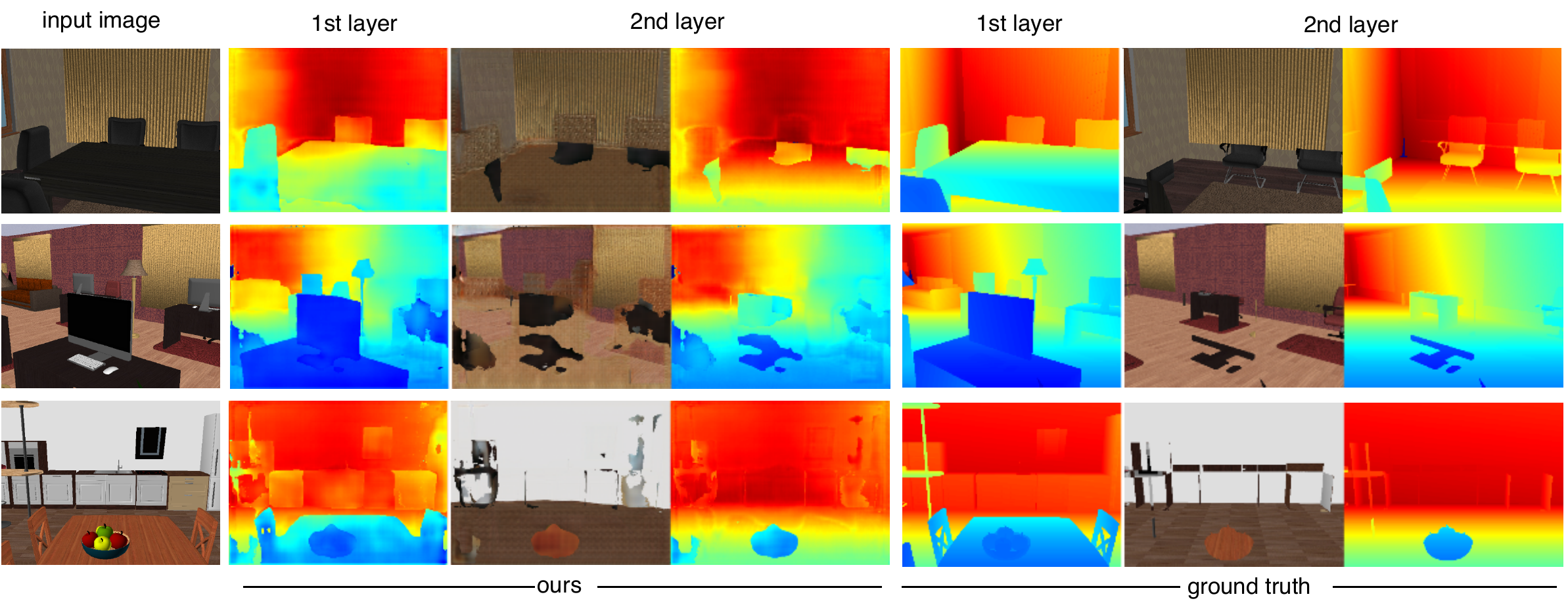}
\caption{\textbf{LDI prediction results on SunCG.} \emph{Left:} The input color image. \emph{Center:} Our predictions for the first two layers, obtained after sorting the object-wise layers. \emph{Right:} Ground truth, as extracted from the mesh-based rendering.}
\label{fig:ldi_suncg}
\end{figure*}

\setlength{\tabcolsep}{15pt}
\begin{table*}[!t]
\small
\begin{center}
\begin{tabular}{l|cc|cc|cc}
\multirow{2}{*}{Method} &  \multicolumn{2}{c}{1st layer depth} & \multicolumn{2}{|c}{2nd layer depth} & \multicolumn{2}{|c}{2nd layer RGB} \\
 & MPE & RMSE & MPE & RMSE  & MPE & RMSE \\
\Xhline{2\arrayrulewidth}
Tulsiani \etal \cite{lsiTulsiani18} & 1.174 & 1.687  & 1.582 & 2.873 & 72.70 & 91.51 \\
Dhamo \etal \cite{Dhamo2018PeekingBO}  & 0.511 & 0.832 & 1.139 & 1.848 & 48.57 & 76.98 \\
\hline
Ours, baseline (w/o class scores)  & 0.551 & 0.879 & 0.687 & 1.120 & 43.97 & 66.51 \\
+ class scores & 0.508 & 0.793 & 0.700 & 1.090 & 44.50 & 65.70 \\
+ $\mathcal{L}_p$ & 0.496 & 0.800 & 0.657 & 1.095 & 43.92 & 66.48\\
+ $ \mathcal{L}_\mathrm{recompose}$ & \textbf{0.473} & \textbf{0.767} & \textbf{0.641} & \textbf{1.071} & \textbf{43.12} & \textbf{65.66}
\end{tabular}
\vspace{5pt}
\caption{\textbf{Evaluation of LDI prediction on SunCG} for the first two layers of depth and the 2nd layer of RGB. We outperform the baselines. The errors are measured for color range $0-255$ and depth in meters.}
\label{table:suncg}
\end{center}
\end{table*}

\textbf{Network B} The layout generator has a U-Net \cite{unet} structure, similar to \cite{pix2pix2016}. The generator G is an architecture with skip connections consisting of seven convolutions and deconvolutions, with a stride of two. The number of filters starts at 64 and is doubled after each convolution. Similarly, the deconvolutions halve the number of feature channels. Encoder and decoder outputs with the same resolution are concatenated. 
The discriminator D consists of 6 convolutions followed by a fully connected layer. Here, the output feature maps contain 64, 128, 256, 512, 512 and 512 channels. All layers in G and D are followed by batch normalization and leaky ReLU. The loss weights are $\lambda_r=100$ and $\lambda_p=25$.

\textbf{Network C} The re-composition block is composed of three $3\times3$ convolutions, each followed by ReLU. 

\textbf{Zero-padding around borders} LDI representations are mostly intended for applications that involve a viewpoint change. As a side effect, the novel image contains empty regions on the borders, when the content was not visible in the original view. Therefore, we add zero padding to our input images before feeding them to the framework. Then, during inference, the the original view is spanned by predicting the originally padded surroundings. For the experiments of this paper, we use padding bands of 16 pixels on the top and bottom and 12 pixels on the left and right borders.

\begin{figure*}[t]
\centering
\includegraphics[width=0.8\linewidth]{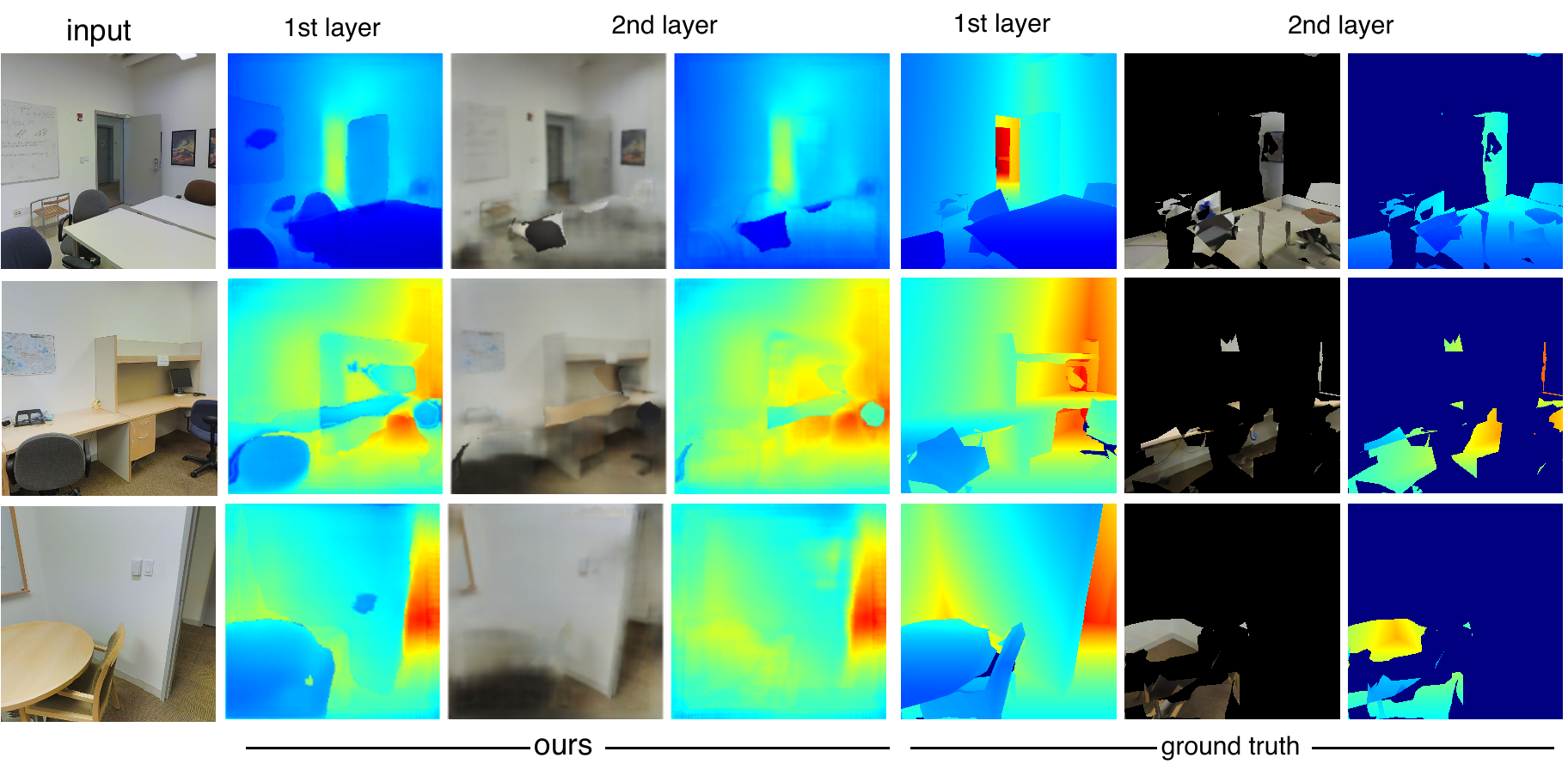}
\caption{\textbf{LDI prediction results on Stanford 2D-3D.} \emph{Left:} The input color image. \emph{Center:} Our predictions for the first two layers, obtained after sorting the object-wise layers. \emph{Right:} Ground truth, as extracted from the mesh-based rendering. Black in the color images and dark blue in the depth maps indicates information holes.} 
\label{fig:ldi_stanford}
\end{figure*}

\begin{table*}[t]
    \centering
    \small
\begin{tabular}{l|cc|cc|cc}
\multirow{2}{*}{Method} &  \multicolumn{2}{c}{1st layer depth} & \multicolumn{2}{|c}{2nd layer depth} & \multicolumn{2}{|c}{2nd layer RGB} \\
 & MPE & RMSE & MPE & RMSE  & MPE & RMSE \\
\Xhline{2\arrayrulewidth}
    Tulsiani \etal \cite{lsiTulsiani18} & 0.805 & 1.088 & 0.954 & 1.230 & 57.42 & 72.65 \\
    Dhamo \etal \cite{Dhamo2018PeekingBO} & \textbf{0.456} & \textbf{0.676} & 0.830 & 1.193 & 42.92 & 55.87 \\
    Ours w/o $\mathcal{L}_\mathrm{recompose}$ & 0.509 & 0.764 & 0.692 & 0.993 & 42.57 & 55.07 \\
    Ours & 0.469 & 0.695 & \textbf{0.688} & \textbf{0.987} & \textbf{42.45} & \textbf{54.92} 
\end{tabular}
\vspace{5pt}
    \caption{\textbf{Evaluation of LDI prediction on Stanford 2D-3D.} LDI predictions for the first two layers of depth and 2nd layer of RGB. The errors are measured for color range $0-255$ and depth in meters.}
    \label{tab:stanford}
\end{table*}

\section{Experiments}

In this section, we present qualitative and quantitative evaluations of our method on two public benchmark datasets: SunCG \cite{song2016ssc} and Stanford 2D-3D \cite{2017arXiv170201105A_armeni}. We merge the output objects into a layered representation, in accordance with the original LDI idea \cite{Shade:1998:LDI:280814.280882} used in related works \cite{Dhamo2018PeekingBO,lsiTulsiani18}. In each pixel, the first layer represents the first visible point along the ray-line, the second layer relates to the next visible surface point and so on. The merging is done by an extended version of MDP, which sorts the depth of the object-wise layers, instead of simply returning the minimum.
In these experiments, we use Mask R-CNN \cite{He2017MaskR} predictions for the mask and class scores as input to our framework. For Stanford 2D-3D, we employ a network trained on the MS-COCO dataset \cite{lin2014microsoft}. For our experiments on SunCG, we finetune the network pre-trained on MS-COCO, using the NYU 40 class categories. As for the input depth predictions, we use the model from Laina \etal \cite{laina2016deeper}, respectively trained on SunCG and Stanford 2D-3D. We chose \cite{He2017MaskR} and \cite{laina2016deeper} as common baselines with available code. We also make this choice for fairness of comparison, since \cite{Dhamo2018PeekingBO} also uses \cite{laina2016deeper} to predict the first layer depth. However, our method is expected to work with various such models.
We evaluate our results on two metrics, Mean Pixel Error (MPE) and Root Mean Square Error (RMSE). The measurements for each layer are done separately, since the difficulty is expected to depend on the layer index. 

\begin{figure*}[t]
\centering
\includegraphics[width=0.65\linewidth]{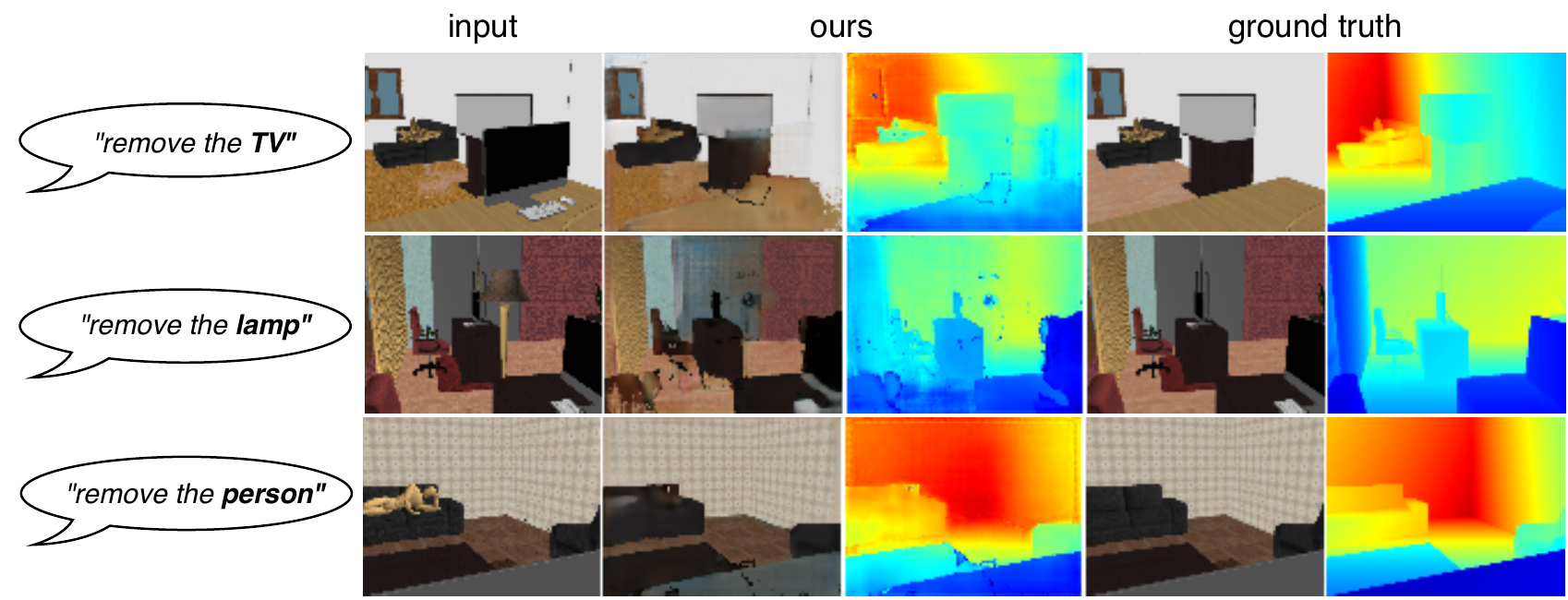}
\caption{\textbf{Illustration of object removal results.} The category labels of the left indicate which object should be removed from the original image. We compare our predicted synthesized images (center) against the ground truth (right). }
\label{fig:object_removal}
\end{figure*}

\begin{figure}[b]
\centering
\includegraphics[width=\linewidth]{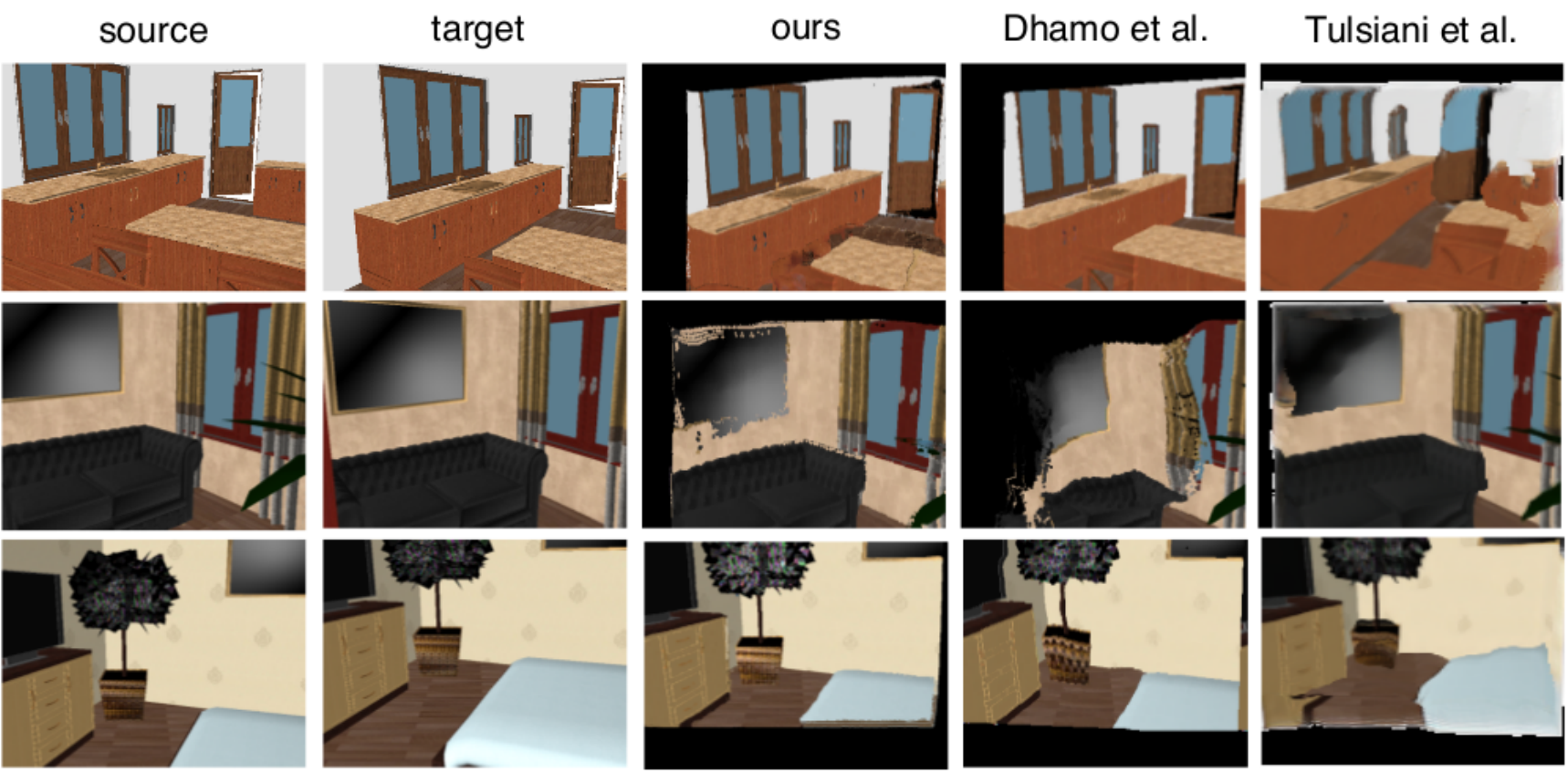}
\caption{\textbf{View synthesis examples.} \emph{Left:} Source image, \ie the input to the proposed method as well as the target image, to be compared with the predictions. \emph{Right:} Predicted novel views, using the LDIs from the proposed method, \cite{Dhamo2018PeekingBO} and \cite{lsiTulsiani18}.}
\label{fig:view_synthesis}
\end{figure}

\setlength{\tabcolsep}{8pt}
\begin{table}[t]
    \centering
    \small
\begin{tabular}{l|ccc}
Method & SSIM $\uparrow$ & MPE $\downarrow$ & RMSE $\downarrow$  \\
\Xhline{2\arrayrulewidth}
    Tulsiani \etal \cite{lsiTulsiani18} & 0.33 & 71.36 & 87.09 \\
    Dhamo \etal \cite{Dhamo2018PeekingBO} & 0.56 & 29.01 & 49.89 \\
    Ours & \textbf{0.65} & \textbf{18.19} & \textbf{34.71}
\end{tabular}
\vspace{5pt}
    \caption{\textbf{View synthesis on SunCG.} The synthesized \textbf{color} images are evaluated in terms of SSIM, MPE and RMSE, in range 0-255.}
    \label{tab:view_synthesis}
\end{table}

\subsection{Layered representation}

\textbf{SunCG} We compare the proposed method against state-of-the-art work in LDI prediction from a single image on the SunCG dataset. The comparison with Dhamo \etal \cite{Dhamo2018PeekingBO} offers insights on the importance of assuming more than one level of occlusion in the scene. In contrast to their hard foreground/background separation, our representation supports more than one level of occlusion (Fig. \ref{fig:ldi_suncg}, second row, desk). On the other end, the comparison with Tulsiani \etal \cite{lsiTulsiani18} confirms the performance gain from a rich supervision. Since current work only evaluates on a two-layer LDI, we utilize our first two layers for the purpose of this experiment. Results on all layers are reported on the supplement. We use the same train and test splits in all these experiments, with 11k images on the training set and 2k on the test set. 
We report results in Table \ref{table:suncg}. We clearly outperform \cite{lsiTulsiani18} and \cite{Dhamo2018PeekingBO} in all metrics, both for color and depth. 
 
Additionally, we wish to point out a qualitative difference between our depth predictions and \cite{lsiTulsiani18,Dhamo2018PeekingBO}. Our method learns the depths instance-wise, therefore it overcomes the common problem of many CNN depth predictors represented by smeared object boundaries (more on the supplement). Sharp object edges are an attractive characteristic in view synthesis, as opposed to smooth edges, which in turn, lead to undesired loss of information during warping. Visual comparisons against these methods are provided in the supplementary material.
Still referring to the results of Table \ref{table:suncg}, one can observe an improvement from adding the class category component, the perceptual loss $\mathcal{L}_p$ as well as our re-composition block, especially for depth.  

\textbf{Stanford 2D-3D}  Given the data limitations, we extracted 14k images with considerable ground truth coverage, from which 13k constructs the train set and 1k is kept for the test set. We follow one of the cross-validation splits suggested in \cite{2017arXiv170201105A_armeni} (area 1,2,3,4,6 vs. area 5a,b). We used the networks pre-trained on SunCG and fine tuned on the Stanford 2D-3D dataset. For this transfer, we convert the MS-COCO classes to NYU 40 to match the categories in our learned SunCG models. We had to disable the GAN loss for the Stanford 2D-3D training, since the network was predicting sparse layouts, trying to mimic a property of the real data. Table \ref{tab:stanford} reports the LDI prediction results. Also here, our method outperforms the baselines for the second layer prediction, which is the main focus of the works. On the first layer, Dhamo \etal results slightly superior, as our problem formulation is more sensitive to holes in the ground truth data and missed detections. However, the results in the synthetic domain encourage further improvement as more complete real datasets become available. In this experiment, we trained Tulsiani \etal on rendered images, as Stanford 2D-3D does not provide raw sequential camera trajectories with high overlap. To ensure that the rendering artifacts do not hinder the consistency between the views (needed for learning through view synthesis), we use rendered images for both the source and the target view (also on test time).  

As shown in Fig. \ref{fig:ldi_stanford}, the ground truth for the second layer appears sparse, although as mentioned in Section \ref{sec:data} we automatically select images where information behind occlusion is available. We only consider the subset of available ground truth pixels for the error measurements.

\subsection{View synthesis}

Generating novel views is a direct application of a LDI representation. Therefore, we perform comparisons in this task. To generate data for this experiment, during the mesh-based rendering, we perturb the camera poses to obtain target frames. For the comparison, we use the same data splits as in the previous experiment. 
We utilize the layered representation learned from \cite{lsiTulsiani18,Dhamo2018PeekingBO} and our proposed method, and apply image-based rendering to synthesize the target views. We employ a simple rendering approach - basically, the first layer is warped first, while the following layers fill the holes left by the first rendering in a sequential manner. The resulting color images are then compared against the ground truth target views. Quantitative results are shown in Table \ref{tab:view_synthesis}. One can clearly observe a better performance of our method, such as alignment with the target, as witnessed by the more accurate color and depth in occluded regions. Fig. \ref{fig:view_synthesis} illustrates how our method preserves the shapes of the front objects while rendering. 

\subsection{Object removal}

Further, we illustrate the application of our method in diminished reality, \ie removal of specific objects from the scene. We take the layered representation as regressed from our framework, where each layer consist of an object or layout. Then, we pass an object category, which we want to be removed from the original image. In this case, we assume fixed input commands of the form "remove \textbf{class}", which satisfies the scope of this work. However, combinations with more advanced natural language processing algorithms would be interesting to explore. Fig. \ref{fig:object_removal} shows a few examples of this application. We observe that our method predicts plausible shapes for partly occluded objects, even when considerable fractions of them are missing. 

\section{Conclusions}

We addressed the timely problem of inferring a layered representation of a scene, where only a single image is known. The proposed model enables a flexible number of output layers, \ie adapts to the scene complexity. We have shown that the method outperforms previous works, especially targeting the occluded regions. Semantic information was incorporated, which has shown to improve the object completion performance. There is still a number of challenges that need to be addressed for further improvement, such as making the textures crisp and realistic, combining the advantages of global and local context with a view to efficiency, or exploiting spatial relations between objects.

\section*{Acknowledgement}

This research has been supported by DFG funds under the project $\#381855581$. We gratefully acknowledge NVIDIA with the donation of a Titan XP GPU used for this research. We would like to thank Nikolas Brasch, Fabian Manhardt and Manuel Nickel for the valuable suggestions and proofreading.

{\small
\bibliographystyle{ieee_fullname}
\bibliography{egbib}
}

\clearpage
\setcounter{page}{1} 

\section*{Supplementary Material}

We present additional evaluation results of our method, such as a visual comparison with \cite{Dhamo2018PeekingBO, lsiTulsiani18} on LDI prediction, ablation of the layout component, intermediate outputs, multi-layer performance, instance segmentation results, a video illustrating our performance in 3D photography, as well as dataset examples.

\subsection*{LDI visual comparison}

\begin{figure*}[t]
\centering
\includegraphics[width=0.99\linewidth]{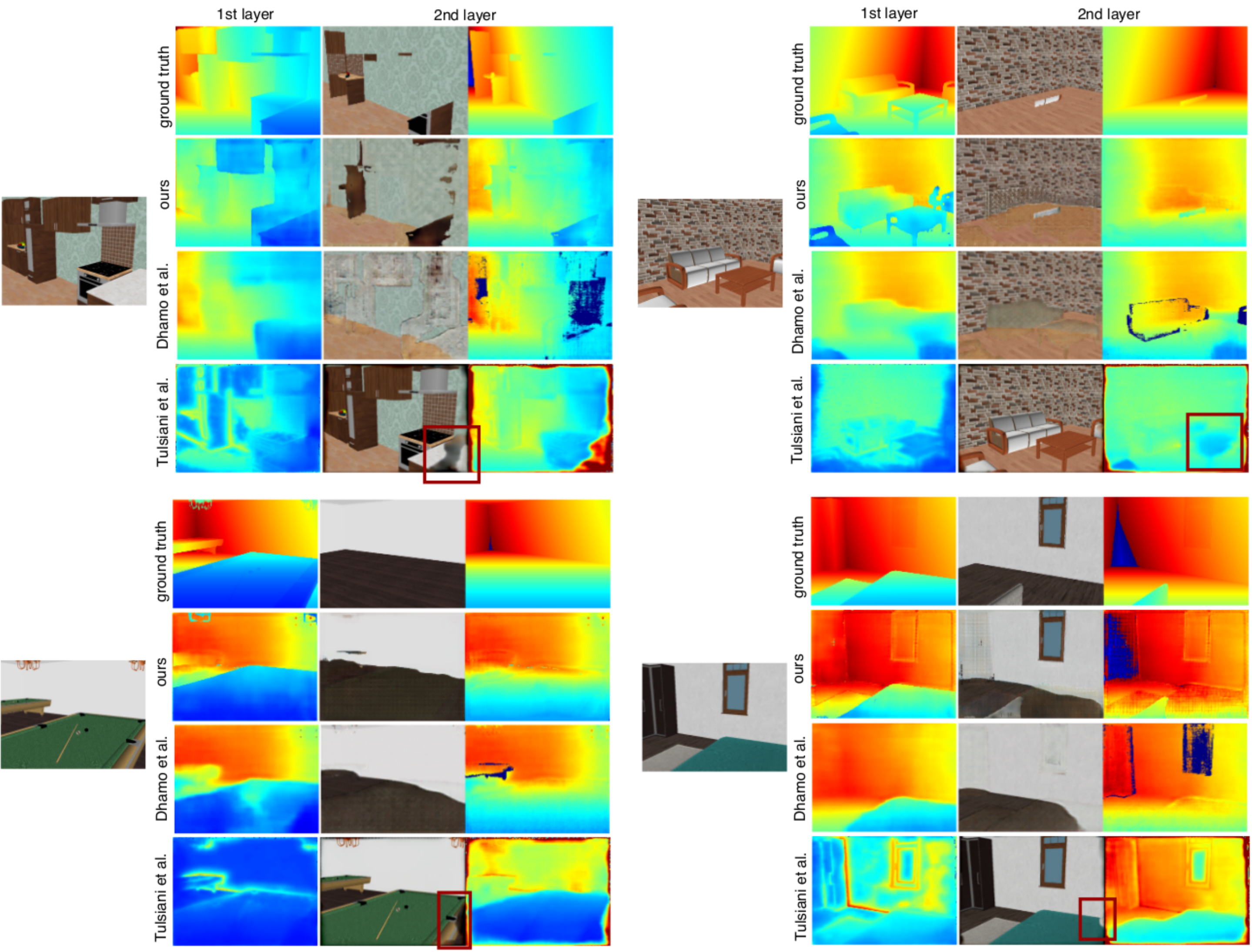}
\caption{\textbf{LDI prediction results on SunCG.} For each example, \emph{Left:} The input color image. \emph{Right:} From top to bottom - ground truth, two-layer predictions of the proposed method, Dhamo \etal \cite{Dhamo2018PeekingBO} and Tulsiani \etal \cite{lsiTulsiani18} for the first two layers.}
\label{fig:ldi_suncg_sup}
\end{figure*}

\begin{figure*}[b]
\centering
\includegraphics[width=0.99\linewidth]{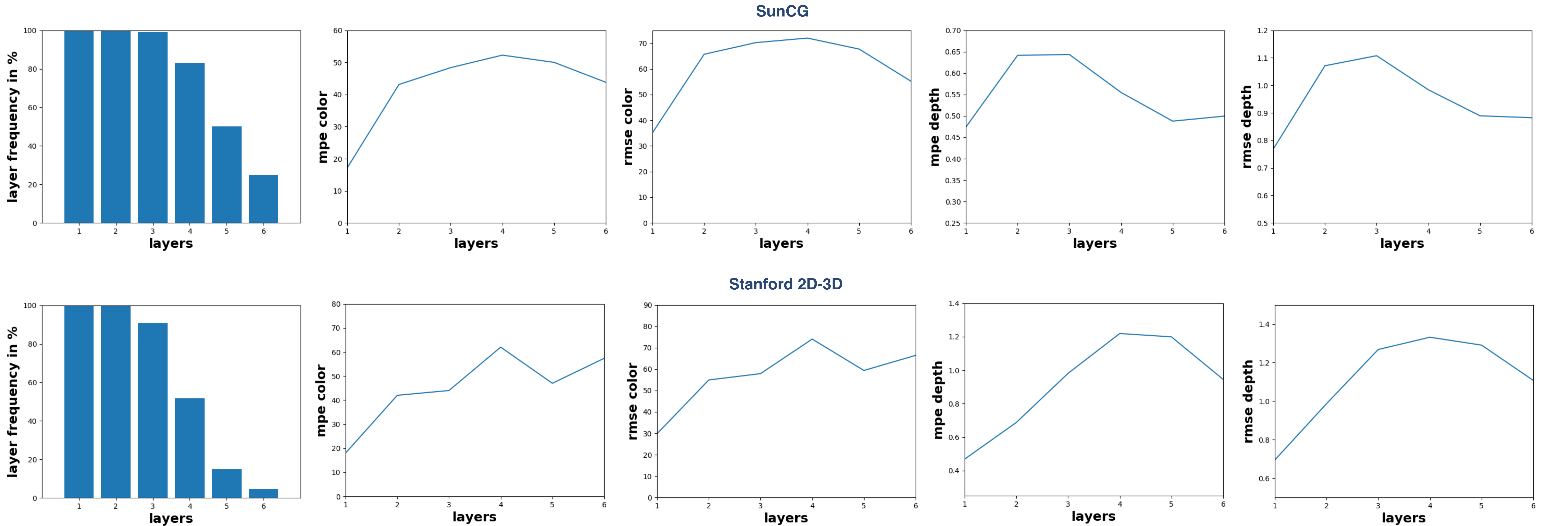}
\caption{\textbf{ Multi-layer evaluation for SunCG (top) and Stanford 2D-3D (bottom).} \emph{Left:} The layer frequency, \ie for layer $l$ the frequency of images that have an $l^{th}$ layer. \emph{Center:} Color MPE and RMSE errors. \emph{Right:} Depth MPE and RMSE errors.}
\label{fig:multi}
\end{figure*}

Fig. \ref{fig:ldi_suncg_sup} illustrates examples of visual comparison between our method, Dhamo \etal \cite{Dhamo2018PeekingBO} and Tulsiani \etal \cite{lsiTulsiani18}. We observe that the method from \cite{lsiTulsiani18} has a rather local diminishing effect, \ie around object borders. In contrast to our method, \cite{Dhamo2018PeekingBO} simply separate the scene in a foreground and a background. For instance, one can see in the upper examples in Fig \ref{fig:ldi_suncg_sup} that the originally occluded regions of foreground object are lost in the layered representation of \cite{Dhamo2018PeekingBO}, while in ours, one can see these parts on the second layer (oven behind furniture, sofa behind table). In the lower left example of Fig. \ref{fig:ldi_suncg_sup}, both methods perform comparably, given that the scene consists of one level of occlusion only.

\subsection*{Ablation of the layout component}

Here, we motivate our design choices for the layout branch (Network B). For this experiment, we compare the layout predictions of our model, against the ground truth layouts. Table \ref{tab:layout} shows that our added loss components improve the performance of the layout prediction, specially for color. In particular, the variant of our model that does not receive a depth prior, leads to considerably less accurate depth. This is an example of performance gain, due to decoupling of a hard task (\ie layout depth prediction from visible color) to simpler tasks (\ie standard depth prediction and RGBD inpainting).

\setlength{\tabcolsep}{6pt}
\begin{table}[!h]
    \centering
    \footnotesize
\begin{tabular}{l|cc|cc}
\multirow{2}{*}{Method} & \multicolumn{2}{c}{color} & \multicolumn{2}{|c}{depth} \\
 & MPE & RMSE & MPE & RMSE \\
\Xhline{2\arrayrulewidth}
    Base, without input depth pred & 21.42 & 42.94 & 0.662 & 1.091  \\
    Base with input depth pred & 22.64 & 42.45 & 0.505 & 0.993 \\
    + adversarial loss & 20.93 & 41.47 & 0.495 & 0.953 \\
    + perceptual loss & \textbf{19.40} & \textbf{39.89} & \textbf{0.482} & \textbf{0.919} \\
\end{tabular}
\vspace{5pt}
    \caption{\textbf{Ablation of the layout prediction (Network B) on the SunCG dataset.} Base refers to the model as introduced in the paper, where only the reconstruction loss is present $\mathcal{L}_r$. The errors are measured for color range $0-255$ and depth in meters.}
    \label{tab:layout}
\end{table}

\subsection*{Layout and object completion}

\begin{figure*}[b]
\centering
\includegraphics[width=0.85\linewidth]{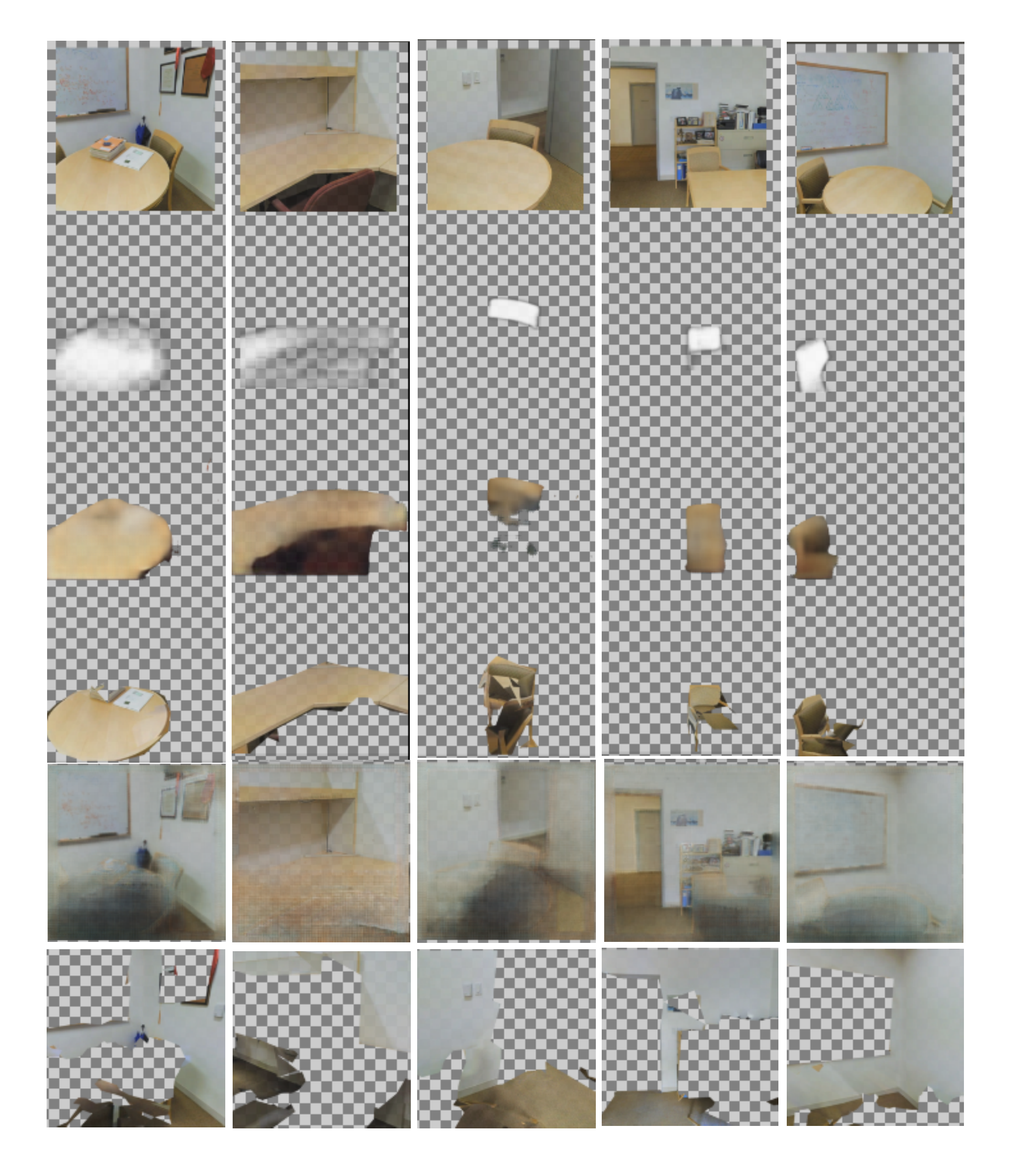}
\caption{\textbf{ RGBA object completion and layout prediction results on Stanford 2D-3D.} Input image, instance examples (top to bottom: mask, prediction, ground truth) as well as layout prediction.}
\label{fig:stanford}
\end{figure*}

\begin{figure*}[b]
\centering
\includegraphics[width=0.99\linewidth]{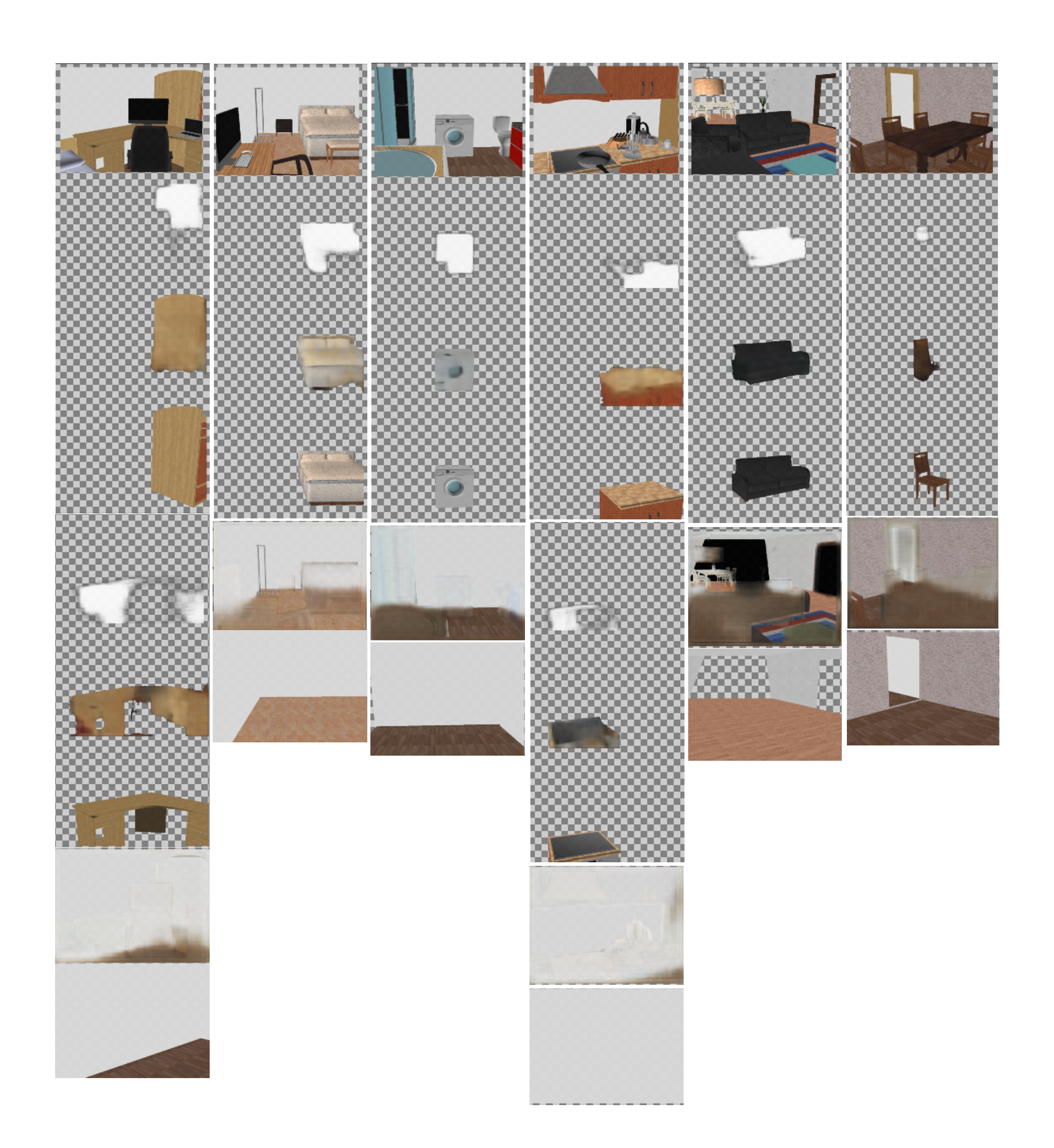}
\caption{\textbf{ RGBA object completion and layout prediction results on SunCG.} Input image, instance examples (top to bottom: mask, prediction, ground truth) as well as layout prediction.}
\label{fig:suncg}
\end{figure*}

In this paragraph we demonstrate an intermediate step of our method, which is object completion and layout prediction. Fig. \ref{fig:stanford} and \ref{fig:suncg} provide examples of the predicted mask probabilities, where opacity indicates confidence. From top to bottom, those are followed by the predictions of our network and ground truth. The two bottom rows visualize the predicted and ground truth layouts. Interestingly, the predictions tent to describe plausible object shape and texture, neglecting the color of front occluding objects. For Stanford 2D-3D, the collected ground truth contains holes, but the network learns from the available examples to regress continuous maps (specially layout).

\subsection*{Accuracy measurements for all layers}

Here we report the error measures of all the predicted layers, for a more thorough insight on the performance of our method. Although not possible to compare against state-of-the-art approaches (restricted to two layers), we find it interesting to see the curve of accuracy as we move from layer to layer. For every layer $l$, whenever there is no novel content (zeros), we migrate the information from the previous layer $l-1$. Then the predicted maps are compared against the ground truth layers, \emph{only in the areas where novel content appears}, \ie ground truth dis-occlusion. This is in accordance with both the LDI representation, as well as the evaluation settings in previous works \cite{Dhamo2018PeekingBO,lsiTulsiani18}. Without this migration, the error values tent to be higher, as it leads to comparing the ground truth with zeros (missing information). Applied to view synthesis, these two settings lead to the same result, as a repetition of previous layers does not lead to novel content on dis-occlusion. 

The results are shown in Fig. \ref{fig:multi}. The frequency plot (left) shows that almost every scene requires three or more LDI layers to be fully represented. As expected, the color and depth errors are the lowest in the first layer, where the level of uncertainty is lower. Further, the errors are roughly comparable in the middle range of layers. Interestingly, we observe a performance increase in the last layers. This is due to the increase of the contribution of the layout component in the composition of later layers. Regressing the box of the scene is an easier problem than completing objects behind occlusion, which makes the layout accuracy higher even behind occlusion. This further supports our choice to decouple the object completion from the layout prediction. 

\subsection*{Instance segmentation}

\begin{figure*}[t]
\centering
\includegraphics[width=0.6\linewidth]{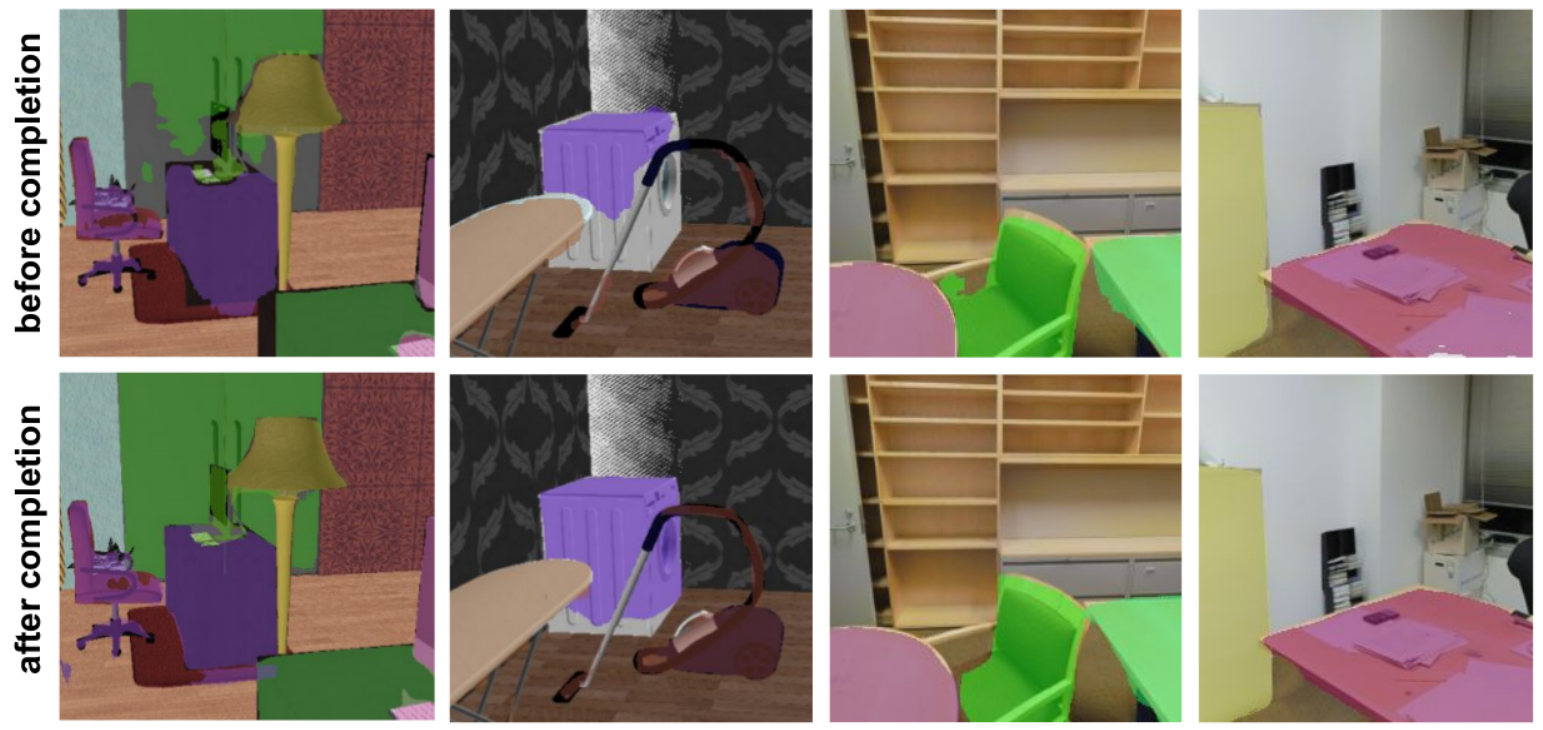}
\caption{\textbf{Visualization of the visible masks.} \emph{Left:} SunCG, \emph{Right:} Stanford 2D-3D. \emph{Top:} Instance masks as predicted from Mask R-CNN, \ie input to our object completion network. \emph{Bottom:} Instance masks using the visible parts of our predicted object extent. We observe that the object completion task inherently refines the visible masks, and aligns them better with the texture borders.} 
\label{fig:mask_vis}
\end{figure*}

We show in Fig. \ref{fig:mask_vis} that our object completion inherently refines the input visible masks.

\subsection*{3D Photography video}
\label{sec:3d}

We demonstrate a 3D Photography video, using our predictions. The frames are from the test set on SunCG and Stanford 2D-3D. We use inverse bilinear interpolation during the image-based rendering, to fill in the holes caused by pixel discretization of the target coordinates. 

\subsection*{Datasets}

\begin{figure*}[t]
\centering
\includegraphics[width=0.95\linewidth]{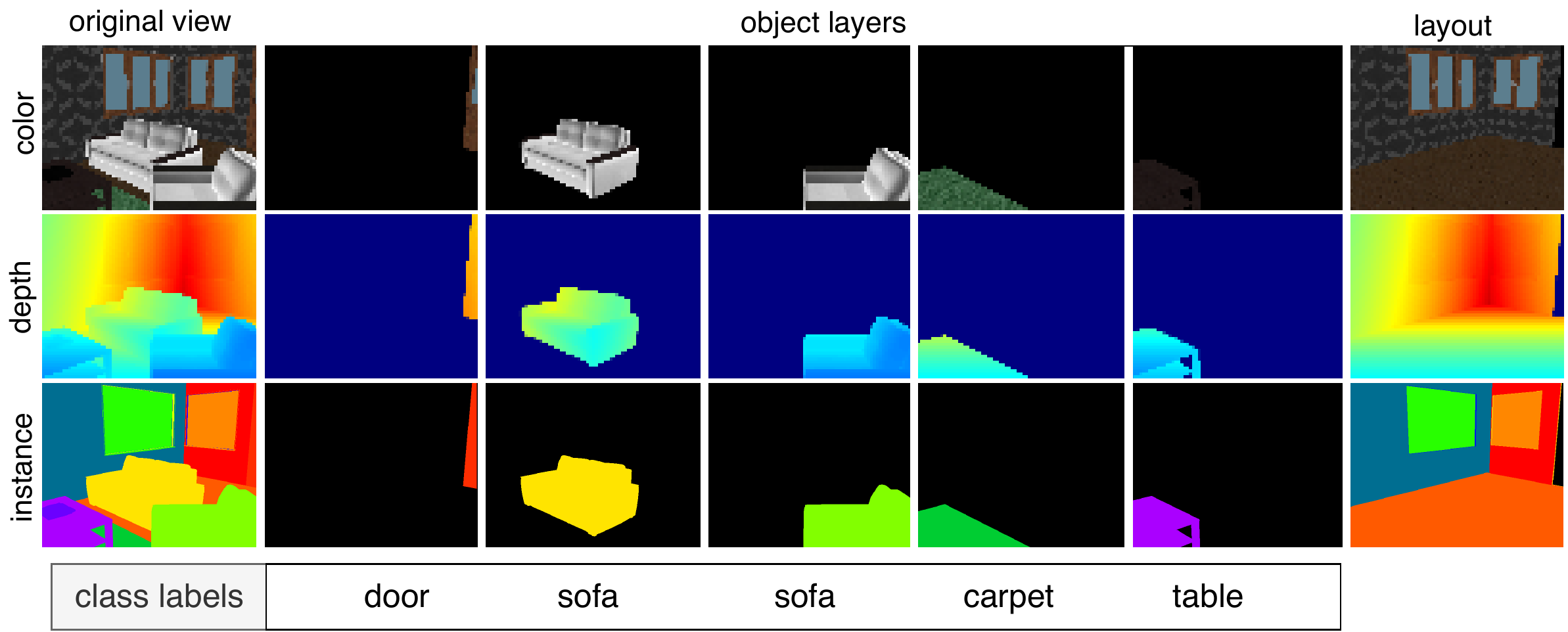}
\caption{\textbf{Illustration of the SunCG dataset.} For every view, we provide the RGBA, depth, instance segmentation and class categories. This applies for the full-image content, object-wise layers as well as the layout. Even though the layout components are merged into a single layer, we keep track of the individual instances, as this can be exploited in future work.} 
\label{fig:dataset_suncg}
\end{figure*}

\begin{figure*}[t]
\centering
\includegraphics[width=0.99\linewidth]{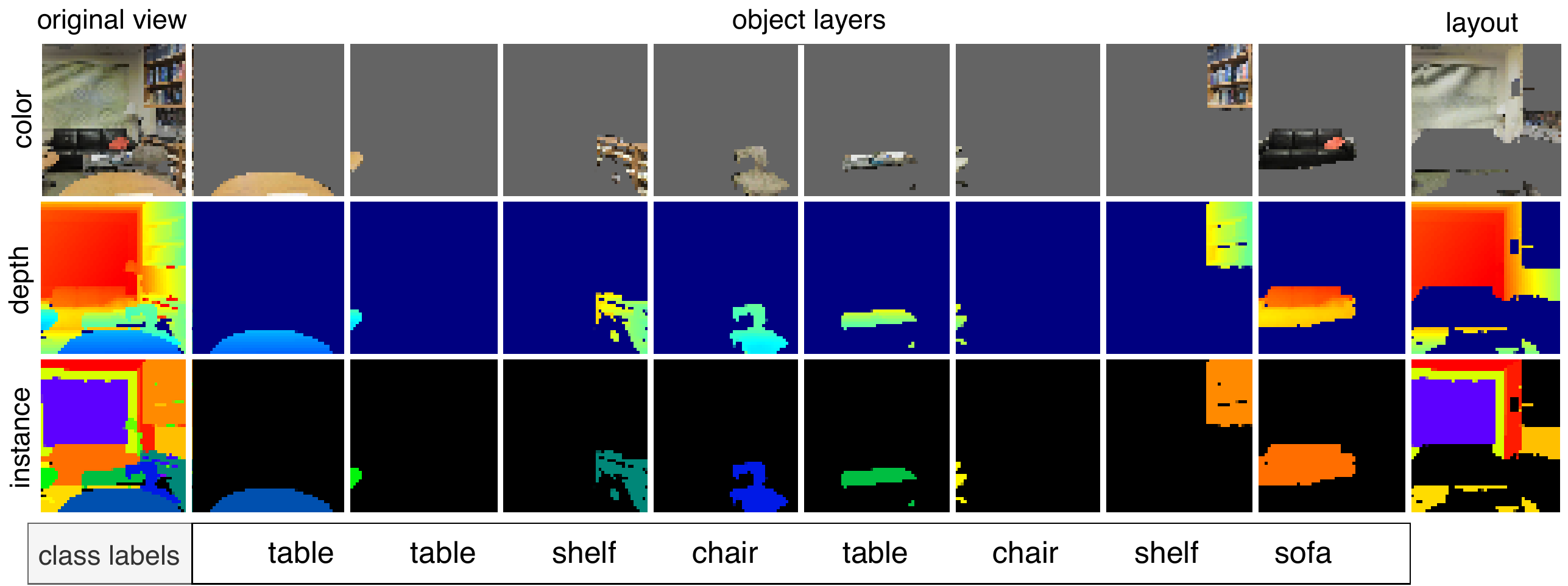}
\caption{\textbf{Illustration of the Stanford 2D-3D dataset.} For every view, we provide the RGBA, depth, instance segmentation and class categories. This applies for the full-image content, object-wise layers as well as the layout. Even though the layout components are merged into a single layer, we keep track of the individual instances, as this can be exploited in future work.} 
\label{fig:dataset_stanford}
\end{figure*}

We show in Fig. \ref{fig:dataset_suncg} and Fig. \ref{fig:dataset_stanford} an example from the automatically generated datasets. 

\clearpage

\subsection*{Training details} 

We train Network A, B and C separately, using the Adam Optimizer with a learning rate of $1 \cdot 10^{-4}$ for Network A, $2 \cdot 10^{-3}$ for Network B and $1 \cdot 10^{-3}$ for Network C. We used a batch size of 4 (resolution $384\times512$) for SunCG and 8 (resolution $256\times256$) for Stanford 2D-3D. 

\subsection*{Failure cases} 

The performance of the proposed method depends on the quality of predicted masks. For instance, if there are repetitions in the detection for a certain object, our algorithm produces two layers. Additionally, objects that are not detected might be lost from the layered representation, especially affecting scenes that contain a considerable amount of objects.

\end{document}